\documentclass[letterpaper]{article} % DO NOT CHANGE THIS
\usepackage{aaai23}  % DO NOT CHANGE THIS
\usepackage{times}  % DO NOT CHANGE THIS
\usepackage{helvet}  % DO NOT CHANGE THIS
\usepackage{courier}  % DO NOT CHANGE THIS
\usepackage[hyphens]{url}  % DO NOT CHANGE THIS
\usepackage{graphicx} % DO NOT CHANGE THIS
\usepackage{natbib}  % DO NOT CHANGE THIS AND DO NOT ADD ANY OPTIONS TO IT
\usepackage{caption} % DO NOT CHANGE THIS AND DO NOT ADD ANY OPTIONS TO IT
\usepackage{algorithm}
\usepackage{algorithmic}
\usepackage{multirow}
\usepackage{mathtools} 
\usepackage{amsmath}
\usepackage{amsfonts} 

\usepackage{newfloat}
\usepackage{listings}
\DeclareCaptionStyle{ruled}{labelfont=normalfont,labelsep=colon,strut=off} % DO NOT CHANGE THIS
\floatstyle{ruled}
\newfloat{listing}{tb}{lst}{}
\floatname{listing}{Listing}
\newcommand{\ourMethod}{StereoDistill}

\title{StereoDistill: Pick the Cream from LiDAR for Distilling \\ Stereo-based 3D Object Detection}
\author{
    Zhe Liu\textsuperscript{\rm 1},
    Xiaoqing Ye\textsuperscript{\rm 2},
    Xiao Tan\textsuperscript{\rm 2},
    Errui Ding\textsuperscript{\rm 2},
    Xiang Bai\textsuperscript{\rm 1}\thanks{Corresponding Author.}
}
\affiliations{
    \textsuperscript{\rm 1}Huazhong University of Science and Technology,
    \textsuperscript{\rm 2}Baidu Inc., China \\
     zheliu1994@hust.edu.cn, yxq@whu.edu.cn, tanxchong@gmail.com, dingerrui@baidu.com, xbai@hust.edu.cn
}

\usepackage{bibentry}
\begin{document}

\maketitle

\begin{abstract}
In this paper, we propose a cross-modal distillation method named \ourMethod{} to narrow the gap between the stereo and LiDAR-based approaches via distilling the stereo detectors from the superior LiDAR model at the response level, which is usually overlooked in 3D object detection distillation.
The key designs of \ourMethod{} are: the X-component Guided Distillation~(XGD) for regression and the Cross-anchor Logit Distillation~(CLD) for classification. In XGD, instead of empirically adopting a threshold to select the high-quality teacher predictions as soft targets, we decompose the predicted 3D box into 
sub-components and retain the corresponding part for distillation if the teacher component pilot is consistent with ground truth to largely boost the number of positive predictions and alleviate the mimicking difficulty of the student model. For CLD, we aggregate the probability distribution of all anchors at the same position  to encourage the highest probability anchor rather than individually distill the distribution at the anchor level. 
Finally, our \ourMethod{} achieves state-of-the-art results for stereo-based 3D detection on the KITTI test benchmark and extensive experiments on KITTI and Argoverse Dataset validate the effectiveness. 
\end{abstract}

\section{Introduction}
3D detectors equipped with LiDAR points~\cite{shi2019pointrcnn,yang20203dssd,deng2020voxel,MV3D,huang2020epnet,liu2022epnet++} for autonomous driving have presented outperforming performance. However, LiDAR sensors usually have a high cost and sensitivity to weather, which limit their application. Alternatively, stereo cameras are capturing increasing interest thanks to their good trade-off in low cost and accuracy. There is still a huge performance gap between stereo-based and cutting-edge LiDAR-based 3D detection methods due to the inaccurate depth estimation by stereo matching. A question naturally arises: can the LiDAR model help to improve the performance of the stereo model?
%In recent years, most leading 3D object detection approaches~\cite{shi2019pointrcnn,yang20203dssd,deng2020voxel,second} for autonomous driving rely heavily on the LiDAR sensor due to its accurate depth information and rich geometric knowledge. Owing to its high cost and sensitivity to weather, LiDAR has its limitation in wide applications. Alternatively, stereo cameras have been capturing extensive interest thanks to their prominence in low cost and robustness. However, there is still a huge performance gap between stereo-based and the cutting-edge LiDAR-based 3D detection methods due to the inaccurate depth estimation by stereo matching. A question naturally arises: can the LiDAR model help to improve the performance of the stereo model?

Knowledge distillation~(KD)~\cite{hinton2015distilling} might be a promising solution for this question, which guides the student model to mimic the knowledge of the teacher model for performance improvement or model compression. The current KD methods of object detection can be mainly classified into the feature-based and response-based streams, in which the former carry out distillation at the feature level~\cite{Zagoruyko2017AT,romero2014fitnets,huang2017like,heo2019knowledge,ye2020monocular,du2020associate} for enforcing the consistency of feature representations between the teacher-student pair whereas the latter adopts the confident prediction from the teacher model as soft targets in addition to the hard ground truth supervision~\cite{yuan2020revisiting,zheng2022LD,dai2021general}. 
However, directly migrating the existing KD methods to LiDAR-to-stereo cross-modal distillation is less effective due to the huge gap between the two extremely different modalities.
The pioneering work LIGA~\cite{guo2021liga} boosted the performance of stereo-based models by applying fine-grained feature-level distillation under the guidance of LiDAR-based models. However, it found little benefit from the response-based distillation due to the erroneous and noisy predictions of the LiDAR teacher.
% tried the cross-modal fine-grained feature distillation to lift the stereo-based detection performance under the guidance of LiDAR with rich geometric features while failing to benefit from the response-based distillation. 

\begin{figure}[t]
	\centering
	\includegraphics[width=0.95\linewidth]{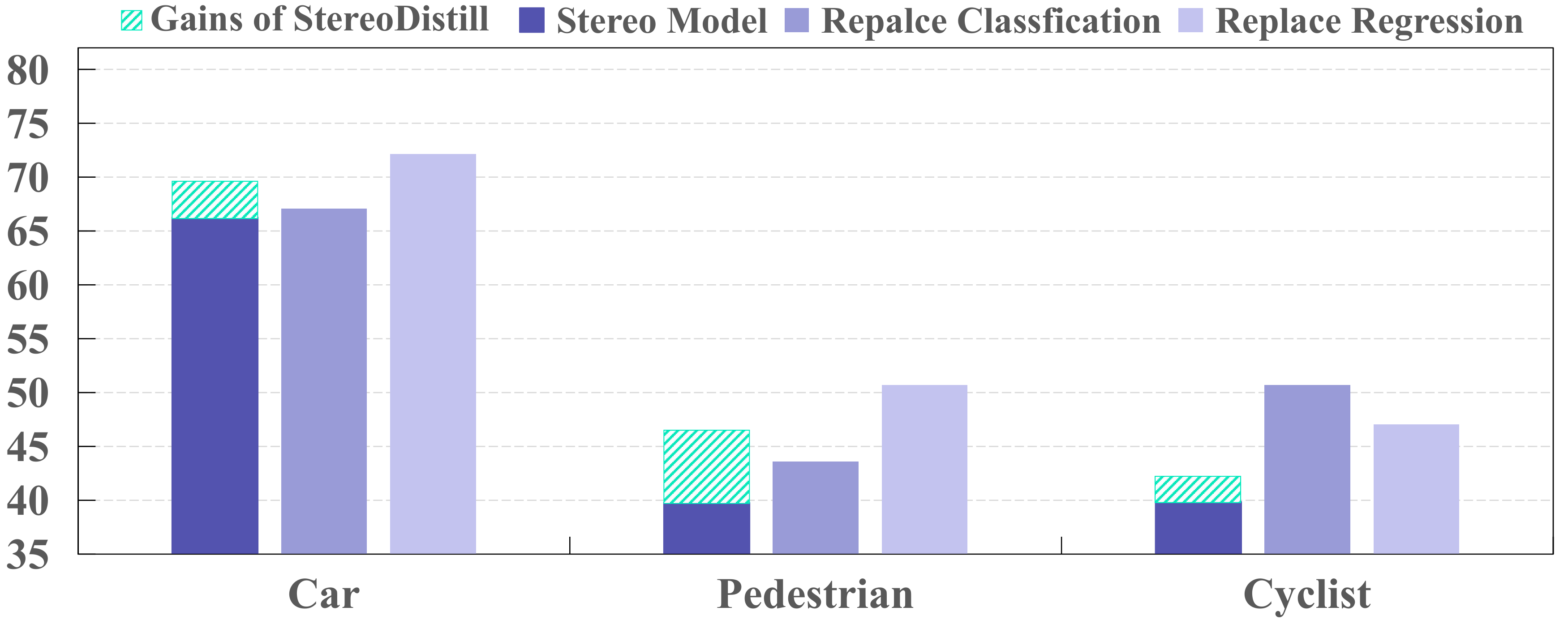}
	\caption{\small{
		3D detection performance (3D mAP) on KITTI \textit{validation} set  of LIGA~\cite{guo2021liga} by replacing the regression and classification results of the stereo model (student) with the teacher LiDAR model SECOND~\cite{second}.
		}}\label{fig:intro}
	% \vspace{-10pt}
\end{figure}

% response distillation
%ye
On the contrary, we argue that the response-level distillation is promising to shrink the gap in the cross-modal domain~(\textit{e.g.}, LiDAR point cloud and binocular images). For illustration, we first obtain the upper bound of the stereo model by replacing its prediction of 3D box regression and classification with the corresponding outputs of the LiDAR model (teacher). As shown in Figure~\ref{fig:intro}, the stereo model with the replaced regression or classification predictions produces impressive results, demonstrating the potential of response-based distillation in the cross-modal domain. 
However, directly applying the vanilla response-level distillation is less effective, either by selecting the high-confident (\cite{yang2022towards}) or high-IoU 3D boxes~(box-level) predicted from the LiDAR model as soft targets~\cite{sun2020distilling}.
The reasons are two-fold: 1) unlike dense 2D images, much fewer high-IoU or high-confident boxes can be adopted as soft labels in a 3D scene due to the high sparsity of LiDAR point cloud; 2) the low-quality boxes discarded by one-size-fits-all thresholds contain underlying beneficial components~(\textit{e.g.}, center, size, or orientation angle) that have been overlooked.

% should be also beneficial to the stereo model as long as one of the components~(\textit{e.g.}, center, size, or orientation angle) decomposed from a 3D box is consistent with the ground-truth.
To tackle the problem, we propose a novel X-component Guided Distillation~(XGD) from the response level. The key idea of XGD is to first decompose a 3D box into sub-X-components ~(X can be center, size, or orientation angle) and retain the beneficial subcomponent as the soft targets if the vector between the teacher's X-component and the student's component is consistent with the vector between the ground truth and the student's, \textit{i.e.}, the two vectors are acute-angled.

% Moreover, the performance improvement through the classification replacement in Figure~\ref{fig:intro} indicates the potential of classification distillation on hard objects~(\textit{e.g.}, Pedestrians and Cyclists). 
%ye
Moreover, we find that only one out of all anchors at the same position can be selected as being responsible for a foreground object in most cases due to the fact that there is usually no overlap among objects in real autonomous driving scenarios, which is different in the 2D domain. Motivated by this observation, we propose a simple and effective Cross-anchor Logit Distillation~(CLD) for classification distillation in our \ourMethod{} to distill by aggregating the confidence distribution of all anchors to a unified distribution so as to highlight the highest probability anchor. 

To summarize, our key contributions are as follows. 
\begin{itemize}
    \item[$\bullet$] We validate that the cross-modal knowledge distillation at the response level can boost the performance of stereo-based 3D object detection. The proposed X-component Guided Distillation~(XGD) for regression avoids the negative effect of erroneous 3D boxes from the LiDAR model by keeping the beneficial X-component as soft targets under the guidance of acute-angled vectors.  
\item[$\bullet$] 
Given the fact that there is no overlap among objects in autonomous driving scenarios, we introduce the simple yet effective Cross-anchor Logit Distillation~(CLD) for classification to aggregate the probability distribution of all anchors at the same position rather than distilling the distribution at anchor level.
% \item[$\bullet$] Without bells and whistles, \ourMethod{} achieves state-of-the-art 3D detection results among stereo-based approaches with no extra cost during inference on the KITTI~\cite{geiger2012we} test benchmark and the large-scale Argoverse dataset~\cite{Argoverse}.
\end{itemize}
\section{Related Works}

\textbf{Stereo-based 3D Object Detection.}
The earlier methods~\cite{stereorcnn,disprcnn,zoomnet} achieve stereo 3D detection based on a strong 2D detector~\cite{ren2015faster,he2017mask}, which does not fully explore the 3D information, leading to suboptimal performance. To introduce more 3D information, ~\cite{pseudolidar,pseudo++,pseudo_e2e} try to convert the estimated depth maps combined with the corresponding image to pseudo point clouds and then can apply the existing LiDAR-based 3D detectors~\cite{second,lang2019pointpillars} to detect 3D boxes. However, directly applying pseudo point clouds for 3D detection might bring erroneous localization due to the limitation of depth estimation, leading to sub-optimal performance.
To tackle this problem, the recent methods DSGN~\cite{chen2020dsgn}, CDN~\cite{garg2020wasserstein}, DSGN++~\cite{chen2022dsgn++} and PLUME~\cite{plume} build cost volume~\cite{flynn2016deepstereo} to encode the implicit 3D geometry features instead of the raw pseudo point representations for 3D object detection. In this paper, we select the prominent DSGN as our stereo model and keep the same configuration with LIGA~\cite{guo2021liga}.

% \begin{figure*}[t!]
% 	\centering\includegraphics[width=\linewidth]{figs/Framework_new_ye.pdf}
% 	\vspace{-15pt}
% 	\caption{The pipeline of our proposed \ourMethod{}. The student and the teacher model take the stereo images and LiDAR points as inputs, respectively. $F_{\{*\}}^{3d}$ and $F_{\{*\}}^{bev}$ are the learned 3D feature and BEV feature map and $*$ denotes the student model ($s$) and the teacher model ($t$). $-D$ is short for Distillation. At the response level, Positive Guidance Distillation~(PGD) and Competitive Logit Distillation~(CLD) are applied on the 3D box regression and classification head, respectively. For feature level, we adopt the similar feature distillation with LIGA~\cite{guo2021liga} with several revisions to obtain better performance.
%     % by considering the attention weight of features and the instance relationship among features. 
% 	% At the feature level, the discriminative feature distillation (DFD) and instance similarity distillation (ISD) are carried out on both the 3D features and BEV features. 
% 	}
% 		\label{framework}
% 	\vspace{-9pt}
% \end{figure*} 

\begin{figure*}[t!]
\begin{center}
% \vspace{-15pt}
  \includegraphics[width=0.9\linewidth]{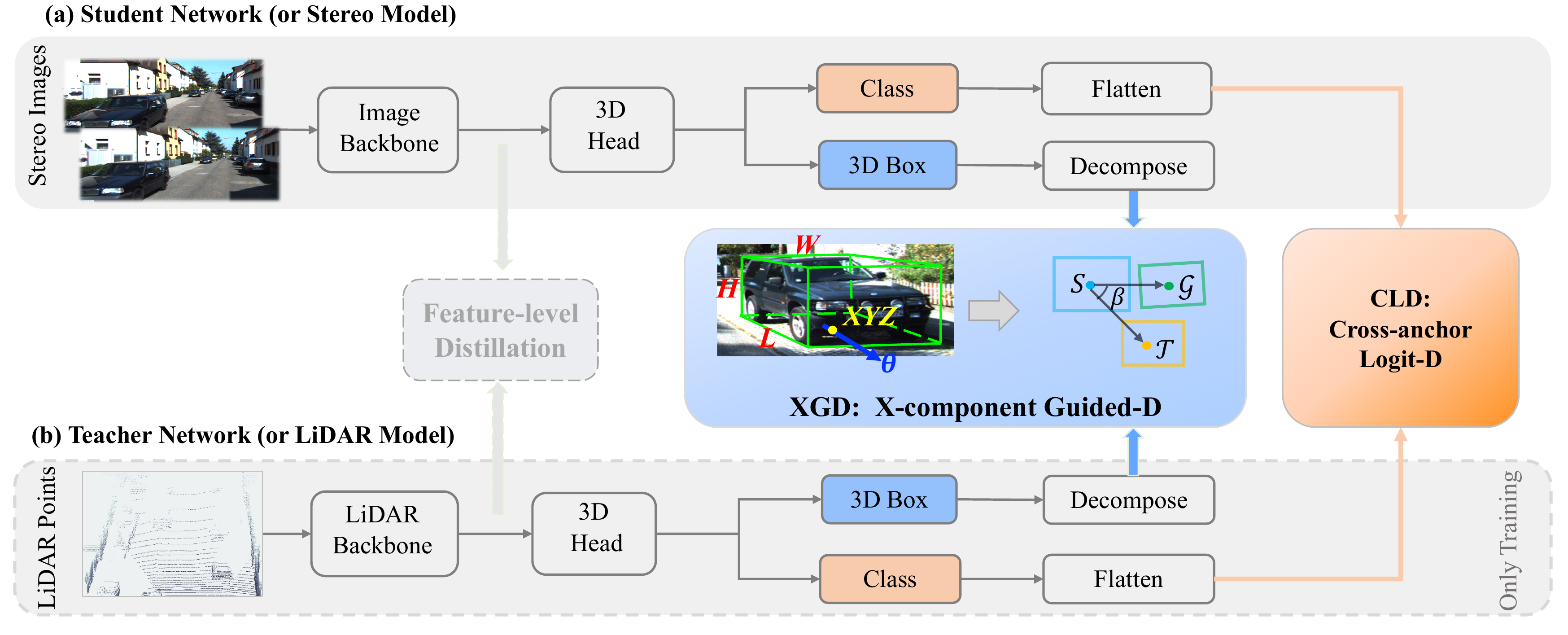}
\end{center}
	\caption{The pipeline of our proposed \ourMethod{} method. The student and the teacher model take the stereo images and LiDAR points as inputs, respectively.  At the response level, X-component Guided Distillation~(XGD) and Cross-anchor Logit Distillation~(CLD) are applied to the 3D box regression and classification head, respectively. In XGD, we decompose the 3D box into sub-components, \textit{i.e.}, size (HWL), center (XYZ) and rotation angle ($\theta$) and keep the components as soft targets if the vectorial angle between teacher-student and GT-student pair is acute. In CLD, we flatten the confidence scores of all anchors falling in the same position and convert them to a unified distribution to highlight the most valuable anchor.
 }
\label{framework}
\vspace{-10pt}
\end{figure*}

\noindent\textbf{LiDAR-based 3D Object Detection.}
Due to the plentiful geometric structure information and accurate depth information from LiDAR sensors, LiDAR-based 3D detectors~\cite{shi2019pointrcnn,second} usually achieve superior performance than the camera-based~\cite{brazil2019m3d,chen2016monocular,li2020rtm3d,simonelli2019disentangling,chen2020dsgn}. 
At present, the mainstream 3D detection methods are divided into two types according to the input data format, including point-based and voxel-based detectors.
The point-based methods~\cite{shi2019pointrcnn,yang20203dssd} usually apply PointNets~\cite{qi2017pointnet,qi2017pointnet++} to deal with this problem of permutation invariance. 
The voxel-based methods~\cite{second,VoxelNet,lang2019pointpillars,liu2019tanet,deng2020voxel} convert the irregular 3D points to the regular voxel grids and employ 2D/3D convolution operation to estimate the final 3D boxes. In this paper, to better align the predictions with the stereo model DSGN~\cite{chen2020dsgn}, we choose the popular voxel-based detector SECOND~\cite{second} as the LiDAR model.

\noindent\textbf{Knowledge Distillation.} 
Knowledge distillation~(KD) is initially proposed by \cite{hinton2015distilling}, which can transfer knowledge from a larger network to a small network to promote the performance or achieve model compression for lightweight devices. Recently, ~\cite{dai2021general,yang2021focal,chen2021distilling,zhang2020improve} achieve feature-based distillation by focusing on the foreground area or considering a weight matrix for the features. LD~\cite{zheng2022LD} implements the difficult problem of localization  distillation  from the response level by converting the regression of bounding boxes to the probability distribution representation. Besides, Cross-modal feature distillation approaches~\cite{chong2022monodistill,guo2021liga} are gaining popularity as a way to take advantage of the complementarity between different modalities. LIGA~\cite{guo2021liga} is the first attempt to explore the fine-grained feature distillation from LiDAR to stereo 3D detector. However, LIGA fails to benefit the stereo model through the response-based distillation due to the erroneous targets from the LiDAR model. In this paper, we propose an X-component Guided Distillation~(XGD) to deal with this problem by retaining the beneficial component which is consistent with ground truth.

\section{Method}
% In autonomous driving, the most existing approaches utilize multi-modal fusion mechanisms to improve the performance of 3D object detection tasks. 
% However, this might cause the entire system to collapse due to the tight coupling of the multiple sensors if one of the sensors fails.
% We argue that cross-modal distillation is an available measure to release the coupling of different modalities. 

% The cross-modal distillation strategy not only can improve the detection performance of single modality but also can avoid the influence of interference among different modalities, which is crucial for the field of autonomous driving. 
% To this end, we \textbf{r}ethink \textbf{e}ffective \textbf{L}iDAR-to-stereo cross-modal \textbf{d}istillation~(\textbf{RELD}) from both the feature and the response aspects for stereo-based 3D object detection. As shown in Figure~\ref{framework}, RELD is composed of a stereo model DSGN~\cite{chen2020dsgn} which is regarded as the student network and a LiDAR model SECOND~\cite{second} which is treated as the teacher network.
% The configures of the stereo model and the LiDAR model are similar with LIGA~\cite{guo2021liga}. For simplicity, note that we do not show two auxiliary network~\cite{guo2021liga} including a 2D detection network and a depth estimation network in the stereo model in Figure~\ref{framework}.
% we keep the similar teacher network SECOND~\cite{second}~(LiDAR model) and student network DSGN~\cite{chen2020dsgn}~(stereo model) with ~\cite{guo2021liga} for fair comparisons. 

In this part, we introduce the proposed cross-modal distillation \ourMethod{}, which consists of the X-component Guided Distillation~(XGD) and Cross-anchor Logit Distillation~(CLD) at the response level.
As shown in Figure~\ref{framework}, we present the pipeline of our \ourMethod{}, which employs a stereo model, DSGN~\cite{chen2020dsgn} for instance, as the student network and a LiDAR model, SECOND~\cite{second} for instance, as the teacher network only for training. 
Although \ourMethod{} contains the feature-level and response-level distillations, our main contribution focuses on the response-level distillation since the effectiveness on the feature-level has been illustrated in LIGA~\cite{guo2021liga}. For the feature level, we mainly revise the feature distillation in LIGA~\cite{guo2021liga} by introducing the attention weight of features~\cite{Zagoruyko2017AT} and the relationship among instance features~\cite{hou2020inter} to further improve the performance, which is regarded as our baseline~(named Improved LIGA). For more details, please refer to our supplementary materials.

For the response-based distillation, however, the predicted boxes~(box-level) from a teacher network inevitably contain false predictions. Therefore, using all predicted boxes directly without any purifying process is likely harmful to the student network and results in a sub-optimal solution~\cite{guo2021liga}. 
To resolve this problem, we propose a novel XGD to preserve the beneficial X-component~(\textit{e.g.}, center, size and angle) decomposed from a box through the proposed positive component updating algorithm.
In addition, we notice that only one out of all anchors at the same position can usually be selected as being responsible for a foreground object in autonomous driving scenarios. Thus, CLD is proposed to highlight the highest probability anchor across all anchors at the same position. 
Next, we introduce the proposed XGD and CLD in detail.

\begin{figure}[h]
\begin{center}
  \includegraphics[width=0.95\linewidth]{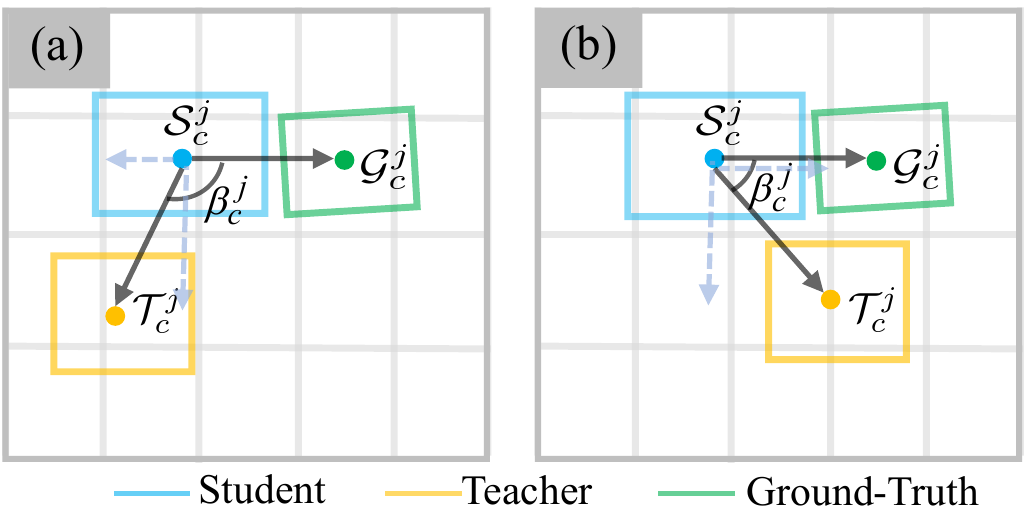}
\end{center}
\caption{Our X-component Guided Distillation (take the center component as an example to illustrate whether the teacher's prediction is beneficial to the student. Case (a) depicts an obtuse angle between the student-to-GT vector and the student-to-teacher vector, showing that the teacher is inconsistent with the GT. Conversely, in Case (b) we observe an acute angle between the two vectors, validating that it is beneficial to be adopted as soft targets to guide the student to regress towards the direction of GT.}
\label{fig:dld}
% \vspace{-10pt}
\end{figure}

\noindent\textbf{X-component Guided Distillation.} 
% In 3D detection tasks, localization regression is critical to the final detection performance. For the stereo network, the localization accuracy especially for distant objects is not so satisfactory due to comparably less depth estimation ability. By contrast, 
As we all know, the LiDAR model has an inherent advantage in localization since the LiDAR sensor can provide more accurate geometrical information and depth information.  
However, the final predictions from the teacher model benefit little from training the stereo network~\cite{guo2021liga}. The main reason is that the erroneous regression of the teacher model may guide the student model to learn in a detrimental direction. 
Although an available solution is to only keep these high-quality boxes for distillation, it brings two flaws. One is that high-quality boxes are too few, resulting in inefficient distillation. The other is that some low-quality discarded boxes can also provide the estimated beneficial component through further decomposing a 3D box into three components~(the center position, the size, and the orientation angle). To be more intuitive, we take the center position as an example and show the harmful and beneficial predicted center position from the teacher model in Figure~\ref{fig:dld} (a) and (b), respectively.

\begin{algorithm}[tb]
\caption{positive component updating}
\label{alg:algorithm}
\textbf{Input}: Boxes of teacher $\mathcal{B}_t=(\mathcal{T}_c, \mathcal{T}_s, \mathcal{T}_o)$, Boxes of student $\mathcal{B}_s=(\mathcal{S}_c, \mathcal{S}_s, \mathcal{S}_o)$, Boxes of GT $\mathcal{B}_g=(\mathcal{G}_c, \mathcal{G}_s, \mathcal{G}_o)$. The number of assigned positive boxes $N_{\mathrm{pos}}$. \\
\textbf{Output}: Updated boxes of teacher $\mathcal{B}_{t*}$.

\begin{algorithmic}[1] %[1] enables line numbers
\STATE Let $\mathcal{T}_{c*},\mathcal{T}_{s*}, \mathcal{T}_{o*}=$[]$,$ []$,$ []
\FOR {$j \in \{1,2,...,N_{\mathrm{pos}}\}$}
\STATE Compute $\cos\beta_c^j$, $\cos\beta_s^j$ and $\cos\beta_o^j$ by the formula~(\ref{formula:dld})
\FOR {$x\in\{c,s,o\}$}
\IF {$\cos\beta_x^j>0$}
\STATE $\mathcal{T}^j_{x*}$  $\leftarrow$ $\mathcal{T}^j_{x}$ 
\ELSE
\STATE $\mathcal{T}^j_{x*}$  $\leftarrow$ $\mathcal{S}^j_{x}$; \# Disable the harmful X-component
\ENDIF
\ENDFOR
\ENDFOR
\\
$\mathcal{B}_{t*}$ = $(\mathcal{T}_{c*}, \mathcal{T}_{s*}, \mathcal{T}_{o*})$
\STATE \textbf{return} $\mathcal{B}_{t*}$
\end{algorithmic}
\end{algorithm}

Motivated by the above observation, we propose a novel component Guided Distillation~(XGD) to tackle this problem. XGD first obtains all the boxes predicted from the assigned positive anchors to keep more valuable 3D boxes. Then XGD selects the `soft boxes’ from these predicted boxes at the X-component level rather than the box level. Specifically, for the predicted $j^\mathrm{th}$ 3D box of the teacher network, we decompose $\mathcal{B}_t^j=(\mathcal{T}_c^j, \mathcal{T}_s^j, \mathcal{T}_o^j)$ into three components, where $\mathcal{T}_c^j=(x_t^j,y_t^j,z_t^j)$, $\mathcal{T}_s^j=(w_t^j,h_t^j,l_t^j)$, and $\mathcal{T}_o^j={\theta}_t^j$, where $\mathcal{T}_c^j$ is the center position of the box along X, Y and Z axes, $\mathcal{T}_s^j$ represents the size including the width, height and length of the 3D box and $\mathcal{T}_o^j$ means the orientation angle of the 3D box. Similarly, we can define the predicted box $\mathcal{B}_s^j=(\mathcal{S}_c^j, \mathcal{S}_s^j, \mathcal{S}_o^j)$ from the student network and the corresponding ground-truth~(GT) assigned boxes of $\mathcal{B}_g^j=(\mathcal{G}_c^j, \mathcal{G}_s^j, \mathcal{G}_o^j)$.
Then, we can judge whether the estimated center $\mathcal{T}_c^j$ from the LiDAR model is beneficial to the stereo model by measuring the cosine value of $\mathcal{T}_c^j - \mathcal{S}_c^j$ and  $\mathcal{G}_c^j - \mathcal{S}_c^j$, which can be formulated as:
\begin{equation}
\label{formula:dld}
\begin{aligned}
\cos{\beta_c^j} = \frac{(\mathcal{T}_c^j - \mathcal{S}_c^j) (\mathcal{G}_c^j - \mathcal{S}_c^j)^T}{\Vert \mathcal{T}_c^j - \mathcal{S}_c^j \Vert_2 \Vert \mathcal{G}_c^j - \mathcal{S}_c^j)^T \Vert_2} \\
\end{aligned}
\end{equation}
Where ${\beta_c^j}$ is the angle between the vector of $\mathcal{T}_c^j - \mathcal{S}_c^j$ and  $\mathcal{G}_c^j - \mathcal{S}_c^j$. Here, when ${\beta_c^j}$ is an acute angle~(or $\cos{\beta_c^j} > 0$), we think the provided center regression $\mathcal{T}_c^j$ from the teacher model can guide the student model regress a more accurate center position. 
Similarly, we can obtain the $\beta_s^j$ and $\beta_o^j$ for the size and angle components following the formulation~(\ref{formula:dld}). 
Then, the final `soft boxes' $\mathcal{B}_{t*}$ is produced by our positive component updating in Algorithm~\ref{alg:algorithm}.
Finally, we employ 3D IoU loss~\cite{zhou2019iou} with rotation as the soft regression term since 3D IoU can comprehensively evaluate the quality of a bounding box. The XGD loss can be computed as:
\begin{equation}
	\begin{aligned}
	  \mathcal{L}_{\mathrm{xgd}} = \sum_{j=1}^{N_\mathrm{pos}} (1 - \mathrm{IoU3D}(\mathcal{B}_s^j, \mathcal{B}_{t*}^j)),
	\end{aligned}
\end{equation}
where $N_\mathrm{pos}$ is the number of the positive anchors in the stereo model and $\mathrm{IoU3D}(\mathcal{B}_s^j, \mathcal{B}_{t*}^j)$ denotes the 3D IoU between $\mathcal{B}_s^j$ and $\mathcal{B}_{t*}^j$.

%%%%%% Table for test set %%%%%%%%% 
\begin{table*}[t!]
\footnotesize
% \vspace{-10pt}
\begin{center}
\begin{tabular}{c|c|ccc|ccc|ccc}
\hline
% header
\multirow{2}{*}{Modality} & \multirow{2}{*}{Method} & 
\multicolumn{3}{c|}{ {Car AP$_{\text{0.7}}$} } & 
\multicolumn{3}{c|}{Pedestrian AP$_{\text{0.5}}$ } & 
\multicolumn{3}{c}{Cyclist AP$_{\text{0.5}}$ } \\ 
\cline{3-11}
& & Easy & {Mod.} & Hard & Easy & Mod. & Hard & Easy & Mod. & Hard \\ 
\hline
% LiDAR results
\multirow{2}{*}{LiDAR}
 & MV3D \cite{MV3D} & 74.97 & 63.63 & 54.00 & -- & -- & - & -- & -- & -- \\
 & SECOND \cite{second} & 83.34 & 72.55 & 65.82 & -- & -- & -- & -- & -- & -- \\
 & AVOD-FPN \cite{AVOD} & 83.07 & 71.76 & 65.73  & 50.46 & 42.27 & 39.04 & 63.76 & 50.55 & 44.93 \\ 
 % & PointPillars \cite{lang2019pointpillars} & 82.58 & 74.31 & 68.99 & 51.45 & 41.92 & 	38.89  & 77.10 & 58.65 & 51.92 \\ % 
 % & PointRCNN \cite{shi2019pointrcnn} & 86.96 & 75.64 & 70.70 & 47.98 & 39.37 & 36.01 & 74.96 & 58.82 & 52.53 \\ %
\hline
% Stereo results
\multirow{10}{*}{Stereo}
& Stereo R-CNN \cite{stereorcnn} & 47.58 & 30.23 & 23.72 & -- & -- & -- & -- & -- & -- \\
& Pseudo-Lidar \cite{pseudolidar} & 54.53 & 34.05 & 28.25 & -- & -- & -- & -- & -- & -- \\
& ZoomNet \cite{zoomnet} & 55.98 & 38.64 & 30.97 & -- & -- & -- & -- & -- & -- \\
& Pseudo-LiDAR++ \cite{pseudo++} & 61.11 & 42.43 & 36.99 & -- & -- & -- & -- & -- & -- \\
& CDN \cite{garg2020wasserstein} & 74.52 & 54.22 & 46.36 & -- & -- & -- & -- & -- & -- \\
& SNVC \cite{li2022stereo} & 78.54 & 61.34  & 54.23 & -- & -- & -- & -- & -- & -- \\ %  % 	%  %
& OC-Stereo \cite{ocstereo} & 55.15 & 37.60 & 30.25 & 24.48 & 17.58 & 15.60 & 29.40 & 16.63 & 14.72 \\
&YOLOStereo3D \cite{liu2021yolostereo3d} & 65.68 & 41.25 & 30.42 & 28.49 & 19.75 & 16.48 & -- & -- & -- \\ %  % 	 % 	     % 	 % 	 %
& Disp-RCNN \cite{disprcnn} & 68.21 & 45.78 & 37.73 & 37.12 & 25.80 & 22.04 & 40.05 & 24.40 & 21.12 \\ %%
& DSGN \cite{chen2020dsgn} & 73.50 & 52.18 & 45.14 & 20.53  & 15.55 & 14.15 & 27.76 & 18.17 & 16.21 \\
& CG-Stereo \cite{cgstereo} & 74.39 & 53.58 & 46.50 & 33.22 & 24.31 & 20.95 & 47.40 & 30.89 & 27.23 \\ % 
& LIGA \cite{guo2021liga} & {81.39} & 64.66 & 57.22 & 40.46 & 30.00 & 27.07 & 54.44  & 36.86  & 32.06 \\
& \textbf{\ourMethod{}}~(\textbf{Ours})  & \textbf{81.66} & \textbf{66.39} & \textbf{57.39} &\textbf{44.12} & \textbf{32.23} & \textbf{28.95} & \textbf{63.96} & \textbf{44.02} & \textbf{39.19} \\ 
\hline
\end{tabular}
\end{center}
\caption{3D Detection results  on the KITTI \textit{test} benchmark. AP$_{\textit{thr}}$ means the threshold value of 3D IoU between the prediction and the ground truth as $\textit{thr}$. `Mod.' is short for Moderate. }
\label{tab:kitti_test_resulst}
\end{table*}

% \begin{table}[ht!]
% \centering
% \footnotesize
% \begin{center}
%     \centering
%     \scalebox{0.95}{\begin{tabular}{c|c|c|c}
%     \hline
%       Method & Easy & Moderate & Hard   \\
%       \hline
%       Disp-RCNN~\shortcite{disprcnn} & 64.29 & 47.73 & 40.11 \\ %  
%       DSGN~\shortcite{chen2020dsgn}& 72.31 & 54.27 & 47.71 \\ %  
%       CG-Stereo~\shortcite{cgstereo} & 76.17 & 57.82 & 54.63 \\ %   
%       LIGA~\shortcite{guo2021liga} &84.92  & 67.07  & 63.80  \\
%       \hline
%       \textbf{Ours} & \textbf{86.17} & \textbf{67.42} & \textbf{64.31} \\
%       \hline
%       \end{tabular}}
%   \end{center}
%       \caption{{3D detection results on \textit{Car} Category with AP$_\text{0.7}$ on KITTI \textit{validation} set  with the evaluation metric of 11 recall positions for fair comparison.}}\label{common_val:kitti_val}
% \end{table}

\begin{table}[ht!]
\centering
\footnotesize
\begin{center}
    \centering
    \scalebox{0.95}{\begin{tabular}{c|c|c|c|c}
    \hline
    \multirow{2}{*}{Method} & 
    \multicolumn{2}{c|}{ {Car AP$_{\text{0.5}}$} } & \multicolumn{2}{c}{Pedestrian AP$_{\text{0.25}}$} \\
    \cline{2-5}
      & 3D  & BEV & 3D  & BEV  \\
      \hline
      SECOND~\shortcite{second} (\textbf{\textit{T}}) & 80.67 & 87.06 & 48.96 & 49.14 \\
      \hline  
      DSGN~\shortcite{chen2020dsgn}  (\textbf{\textit{S}}) & 33.68 & 42.76 & 8.58 & 8.95 \\
      LIGA~\shortcite{guo2021liga} & 34.37 & 45.47 & 10.76 & 11.01 \\
      \hline
      \textbf{Ours}  & \textbf{37.55} & \textbf{46.99} & \textbf{13.70} & \textbf{14.04} \\
      \hline
      \end{tabular}}
  \end{center}
  \caption{{Car and Pedestrian detection results on Argoverse \textit{validation} set with the evaluation metric of 11 recall positions. \textbf{\textit{T}} and \textbf{\textit{S}} denote the teacher and the student.}}\label{common_val:argo_val}
\end{table}

% \subsection{Competitive Logit Distillation}\label{method:acd}

\begin{figure}[t]
\begin{center}
% \vspace{-5pt}
  \includegraphics[width=0.9\linewidth]{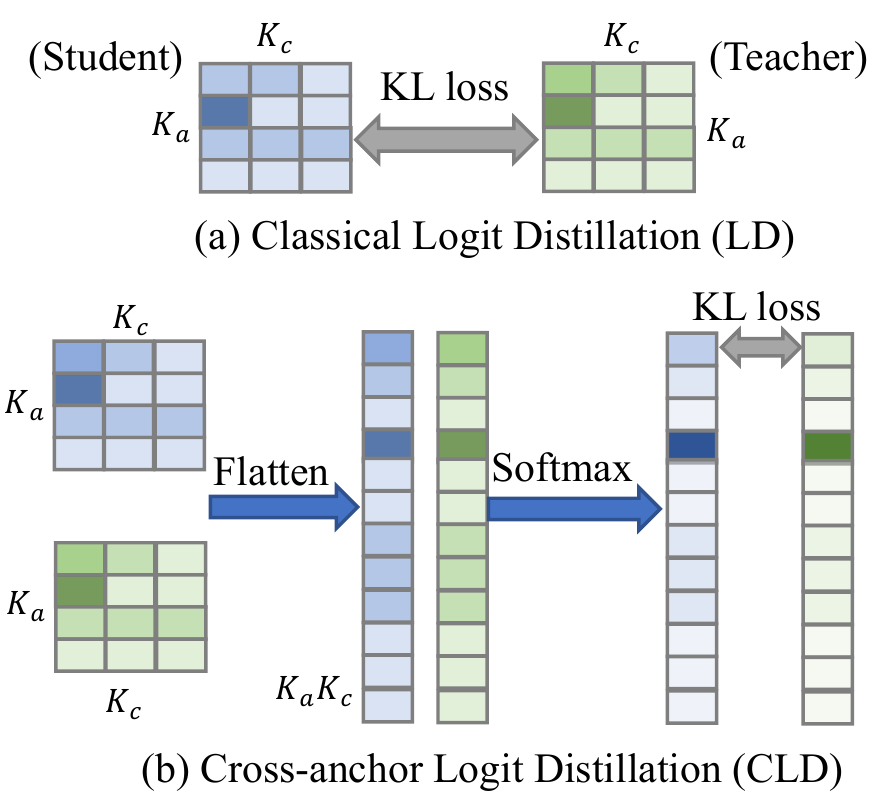}
\end{center}
\caption{The process of the Classical Logit Distillation and our Cross-anchor Logit Distillation. The confidence scores from the student network and the teacher network are marked in blue and green, respectively.  And the darker the color, the higher the confidence.}
\label{fig:ald}
\vspace{-10pt}
\end{figure}

% \textcolor{red}{Competitive$\rightarrow$Cross-anchor}
\noindent\textbf{Cross-anchor Logit Distillation.} Some distillation methods~\cite{chen2017learning,dai2021general,sun2020distilling} via the classification probability usually bring benefits to final results for the 2D detection task, where these distillations are only carried out for positive boxes.
However, our distillation is carried out in all foreground regions since the LiDAR model generates fewer positive 3D samples compared with 2D detection counterparts. Moreover, another distinct characteristic of the 3D detection task against 2D detection lies in the fact that it is rare to find a conflict or overlapping among 3D boxes in autonomous driving scenarios. That is to say, distinct anchors lying in the same position are designed for different objects with different scales and aspect ratios, and hence only one out of these anchors can be selected as being responsible for a foreground object in most cases.  
However, these classical logit distillation approaches~\cite{chen2017learning,dai2021general,sun2020distilling} designed for 2D detection tasks treat anchors separately and do not work well in the 3D detection task, shown in Figure.~\ref{fig:ald} (a).
Given that, we propose a Cross-anchor Logit Distillation~(CLD) approach to highlight the most representative anchor from all anchors in the same position by converting the confidence distribution of each anchor to a unified distribution, whose process is described in Figure.~\ref{fig:ald} (b). Specifically, we first reshape the output confidence map of the teacher network $P_t \in \mathcal{R}^{N_{\mathrm{fore}} \times K_c}$ as $P_t' \in \mathcal{R}^{M_{\mathrm{fore}}\times (K_c K_a)}$, where $N_{\mathrm{fore}} = M_{\mathrm{fore}} K_a$. Here, $M_{\mathrm{fore}}$, $K_a$ and $K_c$ represent the number of all foreground positions, the pre-defined anchors for each position and the categories on the 3D object detection task, respectively. Then, the softmax function 
% with the temperate\footnote{\label{
% temp} {The higher the value of temperature, the smoother the probability distribution of its output, and vice versa.}} of $T$ 
is applied to further normalize the flattened confidence scores $P_t'$ along the dimension of $K_c K_a$ and obtain the unified confidence distribution $P_t^*$ across all anchors at the same position:
\begin{equation}
	\begin{aligned}
	  P_t^* = \mathrm{softmax}(P_t')
	\end{aligned}
\end{equation}
% \begin{equation}
% 	\begin{aligned}
% 	  P_t^* = \mathrm{softmax}(\frac{1}{T} \cdot P_t')
% 	\end{aligned}
% \end{equation}
Similarly, we can get the confidence distribution $P_s^*$ for the student network.
Finally, the CLD loss can be computed by KL divergence:
\begin{equation}
	\begin{aligned}
	  \mathcal{L}_{\mathrm{cld}} = \mathrm{KL}(P_t^*, P_s^*),
	\end{aligned}
\end{equation}
%where $\mathrm{KL}$ means the KL divergence.

% \subsection{Total Loss Function}~\label{method:loss}

\noindent\textbf{Total Loss Function.} We train the stereo model in an end-to-end manner, and the total loss function is as follows:
\begin{equation}
\begin{aligned}
\mathcal{L}_{\mathrm{total}} = \mathcal{L}_{\mathrm{ori}} + \mathcal{L}_{\mathrm{xgd}} + \mathcal{L}_{\mathrm{cld}},
% \beta_1  \mathcal{L}_{\mathrm{vfd}} + \beta_2 \mathcal{L}_{\mathrm{isd}} + \beta_3 \mathcal{L}_{\mathrm{ald}} + \beta_4 \mathcal{L}_{\mathrm{fld}},
\end{aligned}
\label{eq:10}
\end{equation}
where $\mathcal{L}_{\mathrm{ori}}$ denotes the loss function except the feature distillation loss in the LIGA~\cite{guo2021liga}. For training the LiDAR model, we adopt the same loss function with SECOND~\cite{second}.

\section{Experiments}

%%%%%% Table for main ablation studies %%%%%%%%% 
\begin{table*}[tb]
% \vspace{-5pt}
\footnotesize
\begin{center}
\begin{tabular}{c|c|c|ccc|ccc|ccc}
\hline
% header
\multirow{2}{*}{\#} &
\multirow{2}{*}{\ \ \ XGD \ \ \ } &
\multirow{2}{*}{CLD} &
\multicolumn{3}{c|}{{Car AP$_{\text{3D}}$}} & 
\multicolumn{3}{c|}{{Pedestrian AP$_{\text{3D}}$}} &
\multicolumn{3}{c}{{Cyclist AP$_{\text{3D}}$}}  \\
\cline{4-12}
& & & Easy & Mod. & Hard & Easy & Mod. & Hard & Easy & Mod. & Hard  \\
\hline
\hline
\multicolumn{3}{c|}{$\mathrm{SECOND}^\star$~\shortcite{second} (\textit{teacher})} & 
{90.72} & {81.66} & {78.78} & {69.45} & {62.39} & {56.49} & 
{84.97} & {64.06} & {60.21} \\ 
\hline
\multicolumn{3}{c|}{$\mathrm{DSGN}^\dagger$~\shortcite{chen2020dsgn} (\textit{student})}  & 
83.27 & 64.21 & 58.61 & 40.45 & 34.33 & 29.07 & 54.76 & 32.91 & 30.04   \\ % 
\hline
\multicolumn{3}{c|}{$\mathrm{LIGA}^\star$~\shortcite{guo2021liga} (\textit{paper})}   & 
86.84 & 67.71 & 62.02 & 45.54 & 37.80 & 32.09 & 60.00 & 37.31 & 34.25  \\ %         
\hline
\multicolumn{3}{c|}{$\mathrm{LIGA^\dagger}$~\shortcite{guo2021liga} (\textit{reproduced})}  & 
84.32 & 67.14 & 61.93 & 47.16 & 38.97 & 34.09 & 63.98 & 38.49 & 36.01  \\ 

\multicolumn{3}{c|}{Improved $\mathrm{LIGA^\dagger}$~\shortcite{guo2021liga} (\textit{baseline})}  & 86.62 & 67.03 & 61.94 & 47.77 & 40.11 & 35.19 & 65.02 & 40.90 & 37.81 \\
\hline
\hline
\textit{I}  &\checkmark  &--  & 86.78 & 67.65 & 62.43 & 52.89 & 45.37 & 39.40 & 66.25 & 41.38 & 38.27 \\
% \textit{III}   & -- &\checkmark  & \textcolor{red}{86.64} &  67.93 & 60.60 & 48.41 &  40.69 & 35.09 & 66.18 & 40.82 & 37.96 \\
\textit{II}   & -- &\checkmark  & 86.67 &  67.57 & 62.19 & 47.81 &  40.69 & 35.84 & 67.77 & 41.65 & 38.62 \\
\textit{III} & 
\checkmark  &\checkmark  & \textbf{87.57} & \textbf{69.75} & \textbf{62.92} & \textbf{55.19} & \textbf{46.76} & \textbf{40.42} & \textbf{69.43} & \textbf{42.31} & \textbf{39.10}  \\ 
\hline
\multicolumn{3}{c|}{\textbf{Improvement} over \textit{baseline}}   & 
\textbf{+0.95} & \textbf{+2.72} & \textbf{+0.98} & \textbf{+7.42} & \textbf{+6.65} & \textbf{+5.23} & \textbf{+4.41} & \textbf{+1.41}  & \textbf{+1.29} \\  
\hline
\end{tabular}
\end{center}
\caption{Ablation studies for our proposed XGD and CLD on the KITTI \textit{validation} set. $\star$ and $\dagger$ in this table denote the  results  reported in the paper and our reproduced results.  `Mod.' is short for Moderate. } 
\label{table:total_ablation}
\end{table*}

\begin{table*}[h!]
\footnotesize
\begin{center}
\begin{tabular}{c|ccc|ccc|ccc}
\hline
% header
\multirow{2}{*}{Methods} &
\multicolumn{3}{c|}{{Car AP$_{\text{3D}}$}} & 
\multicolumn{3}{c|}{{Pedestrian AP$_{\text{3D}}$}} &
\multicolumn{3}{c}{{Cyclist AP$_{\text{3D}}$}}  \\
\cline{2-10}
&  Easy & Mod. & Hard & Easy & Mod. & Hard & Easy & Mod. & Hard  \\
\hline
\hline
$\mathrm{PointPillars}$~\shortcite{lang2019pointpillars} (\textit{teacher}) & 
88.89 & 78.47 & 75.38 & 62.84 & 56.01 & 51.87 & 82.58 & 62.06 & 58.37 \\ 
\hline
$\mathrm{DSGN}$~\shortcite{chen2020dsgn} (\textit{student})  & 
83.27 & 64.21 & 58.61 & 40.45 & 34.33 & 29.07 & 54.76 & 32.91 & 30.04 \\ 
\hline
\hline
$\mathrm{LIGA}$~\shortcite{guo2021liga}  & 
83.46 & 63.40 & 58.29 & 41.68 & 35.76 & 30.38 & 62.75 &37.28 & 34.25  \\  
 XGD + CLD  & 84.74 &65.49 & 60.13 & 45.98 & 40.18 & 34.75 & 66.67 & 41.00 & 37.95 \\ 
 \textbf{Improvement} &\textbf{+1.28} &\textbf{+2.09} &\textbf{+1.84} &\textbf{+4.30} &\textbf{+4.42} &\textbf{+4.37} &\textbf{+3.92} &\textbf{+3.72} &\textbf{+3.70} \\ 
 \hline
 Improved $\mathrm{LIGA}$~\shortcite{guo2021liga}  & 83.94 & 64.27 & 59.00 & 42.37 & 36.84 & 31.54 & 62.84 & 37.71 & 35.12 \\ 
XGD + CLD  & \textbf{85.24} & \textbf{67.62} & \textbf{60.72} & \textbf{47.81} & \textbf{40.69} & \textbf{34.78} & \textbf{67.10} & \textbf{41.19} & \textbf{38.13}  \\ 
 \textbf{Improvement} &\textbf{+1.30} &\textbf{+3.35} &\textbf{+1.72} &\textbf{+5.44} &\textbf{+3.85} &\textbf{+3.24} &\textbf{+4.26} &\textbf{+3.48} &\textbf{+3.01} \\
\hline
\end{tabular}
\end{center}
\caption{Generality of our \ourMethod{}~(XGD + CLD) on the KITTI \textit{validation} set. We select the popular LiDAR model  PointPillars~\cite{lang2019pointpillars} as the teacher model. `Mod.' is short for Moderate.}
\label{table:pillar_teacher}
% \vspace{-8pt}
\end{table*}

\subsection{Experimental Datasets and Evaluation Metrics} 
\textbf{KITTI.} The KITTI dataset~\cite{geiger2012we} includes 7,481 training and 7,518 testing stereo
image pairs with the corresponding LiDAR point clouds. We further split the training data into \textit{training} set with 3712 samples and a \textit{validation} set with 3769 samples following ~\cite{chen2020dsgn,qi2018frustum,shi2019pointrcnn}.  The  evaluation metric~\cite{simonelli2019disentangling} adopts the mean Average Precision~(mAP) with 40 recall positions. If not specified, the metric of all results in the following tables uses 40 recall positions. 
In this paper, we evaluate our method on the \textit{validation} set and the \textit{test} benchmark for three categories of Cars, Pedestrians and Cyclists under three difficulty levels~(\textit{e.g.}, Easy, Moderate, and Hard).

\noindent\textbf{Argoverse.} The Argoverse dataset~\cite{Argoverse} contains 3D detection and tracking annotations from 113 scenes. Different from the Waymo~\cite{sun2020scalability} and Nuscnes~\cite{caesar2020nuscenes} datasets, Argoverse provides stereo image pairs, which can be adopted to verify the generality of our method. For convenience, we convert the Argoverse dataset to the format of KITTI following ~\cite{wang2020train} and obtain a \textit{training} set with 13122 samples and a \textit{validation} set with  5015 samples. We adopt the same evaluation metric with KITTI. 

\subsection{Implementation Details}
For the stereo model DSDN~\cite{chen2020dsgn} and the LiDAR model SECOND~\cite{second}, we use the same network structure with LIGA~\cite{guo2021liga} for fair comparisons. The stereo model is trained on 4 NVIDIA V100 GPUs with a batch size of 4 and is optimized by Adaptive Momentum
Estimation~(Adam)~\cite{kingma2014adam} with the initial learning rate, weight
decay, and momentum factor set to 0.003, 0.01, and 0.9,
respectively. Random horizontal flipping is adopted for data augmentation. For both the KITTI dataset and the Argoverse dataset, we employ the range of the detection area to [-30, 30], [-1, 3], [2, 59.6] meters along the X (right), Y (down), Z (forward) axis in the camera coordinate. The voxel size of the LiDAR model is (0.2, 0.2, 0.2) meters and the volume size of the stereo model is (0.05, 0.1, 0.05) meters. All experiments are conducted on a single model for multiple categories. For more details, please refer to our supplementary materials.

\subsection{Comparisons with State-of-the-art Methods}

\textbf{Evaluation on KITTI.} In Table~\ref{tab:kitti_test_resulst}, we present quantitative comparison with the leading stereo-based 3D detectors and several popular LiDAR-based 3D detectors on the KITTI \textit{test} benchmark. Our method outperforms the SOTA model LIGA~\cite{guo2021liga} with 1.73\%, 2.23\% and 7.16\% 3D mAP on Cars, Pedestrian and Cyclists at the moderate difficulty level, without introducing any extra cost during inference. Our \ourMethod{} even surpasses the LiDAR-based method MV3D~\cite{MV3D} with 3D mAP of 2.76\% on Cars. These superior results demonstrate the effectiveness of our \ourMethod{}. 
For visualization, please refer to our supplementary materials.
%Besides, we provide 3D detection results of our \ourMethod{} on the KITTI \textit{validation} set in Table~\ref{common_val:kitti_val}, which demonstrates consistent performance improvement over the stereo-based DSGN~\cite{chen2020dsgn}, CG-Stereo~\cite{cgstereo} and LIGA.

\noindent\textbf{Evaluation on Argoverse.} 
To further verify the generality of our proposed method, we conduct experiments on the Argoverse dataset. For fair comparisons, we adopt the same network of student and teacher models with LIGA~\cite{guo2021liga} and also re-implement LIGA under the same setting on the Argoverse dataset. 
In Table~\ref{common_val:argo_val}, we present the results with the 3D IoU thresholds of 0.5 and 0.25 for both the BEV and 3D detection on moderate Cars and Pedestrians. Our method exceeds LIGA with 3D mAPs of 3.18\% and 2.94\% and BEV mAPs of 1.52\% and 3.03\%  on  Cars and Pedestrians, which validates the generality of our method.

\subsection{Ablation Studies}

\textbf{Ablation Studies on \ourMethod{}.} In this part, we verify the effectiveness of 
the proposed compositions, including the XGD for regression, CLD for classification, and their combinations in \ourMethod{}. The baseline model is our improved LIGA~\cite{guo2021liga} by further enhancing the feature distillation~(refer to our supplementary materials). In Table~\ref{table:total_ablation}, by comparing  (\textit{I}), (\textit{III}) with the baseline model, the proposed XGD and CLD bring consistent improvements over the baseline on all difficulty levels
for three categories, which demonstrates their effectiveness. Note that the proposed XGD greatly boosts the detection performance on small objects~(\textit{e.g.}, Pedestrians), which requires more accurate regression. It illustrates that the X-component guided distillation can indeed transfer superior location awareness from the LiDAR model to the stereo model so as to obtain better performance.
Integrating these two ingredients, \ourMethod{} in (\textit{III}) outperforms the baseline with remarkable margins of 2.72\%, 6.65\% and 1.41\% 3D mAP on the moderate Cars, Pedestrians and Cyclists, respectively.

\noindent\textbf{Generality of \ourMethod{}.} 
To verify the generality of \ourMethod{}, we replace the common teacher network SECOND~\cite{second} with the other popular 3D detector Pointpillars~\cite{lang2019pointpillars}.
In Table ~\ref{table:pillar_teacher}, we provide two baseline settings: the original LIGA~\cite{guo2021liga} in \textit{Line 4} and the improved LIGA in \textit{Line 7}. 
Not surprisingly, our \ourMethod{} yields obvious performance gains on all difficulty levels for three categories, which further the superiority and generality of our proposed XGD and CLD.

% Comparing \textit{IV} and baseline, the proposed DFD boosts the results of Pedestrians, which indicates that DFD pays more attention to hard objects~(Pedestrians) by considering the discriminative features. 
% As shown in (\textit{I} vs baseline), (\textit{V} vs \textit{II}), (\textit{IX} vs \textit{VI}) and (\textit{XI} vs \textit{X}), the novel FLD attains significant improvements on moderate Pedestrians with 3D mAPs of 4.67\%, 6.49\%, 5.54\% and 6.07\%, 
% demonstrating the superiority of FLD by only retaining the high-fidelity localization information. Moreover, the combination of FLD and CLD in (\textit{V}) is more powerful than the combination  of the proposed feature-based distillation in (\textit{VI}), which proves the necessity of response-based distillation. As described in (\textit{VII} vs \textit{V}) and (\textit{VIII} vs \textit{V}), further improvements are obtained through  
% integrating the ISD or DFD into the response-based distillation, validating that  better feature representation can effectively assist the subsequent response-based distillation. Finally, by integrating all the ingredients, our \ourMethod{} in (\textit{XI}) outperforms the baseline with remarkable margins of 2.61\%, 7.79\%, 3.82\% 3D mAP on the moderate Cars, Pedestrians and Cyclists, respectively.

\begin{table}[t!]
% \vspace{-10pt}
  \centering
  \footnotesize
  %\scriptsize
  \begin{center}
    \centering
    \scalebox{0.9}{\begin{tabular}{c|c|ccc}
    \hline
     \# & Methods &Cars & Pedestrians & Cyclists  \\
      \hline
      \textit{I} & Baseline  &67.57	&40.69	&41.65  \\
      \hline
      \textit{II} &XGD-Center &68.22	&46.33	&42.19  \\
      \textit{III} &XGD-Size &68.02 &42.46	&40.71  \\
      \textit{IV} &XGD-Angle &67.84	&44.11	&41.18  \\
      \textit{V} & \textbf{XGD} &\textbf{69.75} &\textbf{46.76} & \textbf{42.31}  \\
      \hline
      \textit{VI} & High-quality boxes &67.72	 &43.65 & 42.11  \\ %&67.72	&43.65	&42.11
      \textit{VII} & \textbf{Positive anchors~(Ours)} &\textbf{69.75} &\textbf{46.76} & \textbf{42.31}  \\
      \hline
      \end{tabular}}
  \end{center}
  \caption{Ablation studies for XGD.  The results are evaluated with 3D mAP on the moderate difficulty level for Cars, Pedestrians and Cyclists, respectively. XGD-* means the manner of only adopting * for computing XGD loss.}
  \label{sub_abl:XGD}
\end{table}

\begin{table}[t!]
  \centering
  \footnotesize
  \begin{center}
    \centering
    \scalebox{0.9}{\begin{tabular}{c|c|ccc}
    \hline
      \# &Methods &Cars & Pedestrians & Cyclists  \\
      \hline
      \textit{I} & Baseline &67.65	&45.37	&41.38  \\
     \hline
      \textit{II} & Positive Anchors &67.66	&45.71	&\textbf{42.32} \\
      \textit{III} & \textbf{All Foregrounds~(Ours)} &\textbf{69.75} &\textbf{46.76} & {42.31}  \\
      \hline
      \textit{IV} & Classical~\cite{chen2017learning}  &67.74	&46.18	&40.69  \\
      \textit{V} & \textbf{CLD~(Ours)} &\textbf{69.75} &\textbf{46.76} & \textbf{42.31} \\
      \hline
      \end{tabular}}
  \end{center}
   \caption{{Ablation studies for CLD. The results are evaluated with 3D mAP on the moderate  Cars, Pedestrians and Cyclists, respectively.}}
   \label{sub_abl:CLD}
   % \vspace{-5pt}
\end{table}

\noindent\textbf{Analysis of XGD.} In Table~\ref{sub_abl:XGD}, we conduct extensive ablation studies to analyze the effectiveness of XGD. The baseline~(\textit{I}) is our \ourMethod{} without XGD loss. Then, we decompose the regression of 3D boxes into three components including the center position, the size, and the orientation angle to analyze the effect of each component on XGD. We observe that XGD with only the center position~(\textit{II}) exhibits the most competitive performance of the three~(\textit{II}, \textit{III}, \textit{IV}), which illustrates the positive guidance of the center position is the crucial component to help the student model to acquire more beneficial localization information.
Finally, combined with these three components, XGD~(\textit{V}) exceeds the baseline~(\textit{I}) with 3D mAP of 2.08\%, 6.07\% and 0.66\% on moderate Cars, Pedestrians, and Cyclists, demonstrating the superiority of XGD. Moreover, compared with retaining high-quality boxes in (\textit{VI}) whose confidence scores are greater than 0.3~(a proper threshold), our manner of adopting all positive anchors in (\textit{VII}) has obvious gains on Cars and Pedestrians. In the real scene, these two categories usually occupy a much larger number than Cyclists, which means that there may be more false positives on Cars and Pedestrians. 
This demonstrates that our XGD provides a reasonable workaround to deal with some low-quality boxes by retaining the beneficial X-component but discarding the harmful X-component decomposed from a 3D box.

\noindent\textbf{Effectiveness of CLD.}  In Table~\ref{sub_abl:CLD}, we first present the results of distilling the confidence distribution from two alternative regions. The way based on the foreground~(\textit{III}) exceeds the approach by only considering the positions from positive anchors~(\textit{II}) on average, which illustrates the importance of introducing the useful foreground positions beyond the positions of the positive anchors. Furthermore,  we provide a classical logit distillation~(\textit{IV})~\cite{sun2020distilling} as a comparison, which individually treats the confidence distribution of each anchor from each position. It can be observed that our CLD boosts the performance with mAP of 2.01\%, 0.58\% and 1.62\% on Cars, Pedestrian and Cyclists, clearly demonstrating the effectiveness of underlining the confidence distribution for the best competitive anchor from all anchors in a position.

% \begin{figure}[t]
% 	\centering
%  % \vspace{-10pt}
%     \includegraphics[width=0.92\linewidth]{figs/visualization.pdf}
% 	\caption{{The visualization results on the KITTI \textit{validation} set.  We localize the predicted objects from LIGA and our \ourMethod{} with the yellow 3D bounding boxes. The ground truth is colored in green. Besides, we highlight some objects with the red rectangle for comparison.
% 		}} \label{vis:kitti_comparsion}
%   \vspace{-5pt}
% \end{figure} 

% \noindent\textbf{Analysis of Visualization.}
% Figure~\ref{vis:kitti_comparsion} presents the qualitative comparison of the representative approach LIGA~\cite{guo2021liga} and our \ourMethod{} on the KITTI \textit{validation} set. In the second row, LIGA fails to localize a  partially occluded Car correctly. However, our method can detect the object accurately, which benefits from leveraging the beneficial localization information from the LiDAR model through response-based distillation. 
% For more visualization, please refer to our supplementary materials.

\noindent\textbf{Comparison of Different Distillation.} 
In Table~\ref{sub_abl:cmp}, we individually present the comparisons for adopting the feature-based in LIGA~\cite{guo2021liga} or  response-based distillation in \ourMethod{} based on the stereo model DSGN~\cite{chen2020dsgn}. 
First, these two distillations can consistently boost performance over the baseline DSGN~\cite{chen2020dsgn}. Then,
the proposed response-based distillation of our XGD and CLD in \ourMethod{} even outperforms the feature-based distillation in LIGA~\cite{guo2021liga} with 3D mAP of 1.92\% on average  under the same setting, which further demonstrates the effectiveness of the proposed XGD and CLD. 
Moreover, combined with the response-based distillation and the feature-based distillation, our \ourMethod{} produces superior performance over the baseline model DSGN with a 3D mAP of 10.34\% on average. 
This reveals that superior feature representations often lead to better responses, which in turn can further facilitate feature learning. 

\begin{table}[t!]
  \centering
  \footnotesize
  \begin{center}
    \scalebox{0.9}{\begin{tabular}{c|ccc|c}
    \hline
       Methods &Cars & Pedestrians & Cyclists & Mean  \\
      \hline
    %    DSGN~(official) &52.29	&31.42	&23.16  & 35.62 \\ 
    % \hline
    % \hline
       DSGN~(LIGA wo FD) &63.32	&34.23	&30.26 & 42.60 \\  
       \hline
       \hline
       + FD  &67.14	&38.97	&{38.49}  & 48.20 \\
       \textbf{Improvement} &+3.82 &+4.74 &+8.23 &+5.60 \\
       \hline
       + our RD  &{67.42}	&{45.28}	&37.66  & {50.12} \\
       \textbf{Improvement} &+4.10 &+11.05 &+7.4 &+7.52 \\
    \hline
    \hline
       \ourMethod{}  &69.75	&46.76 &42.31  & 52.94 \\ 
       \textbf{Improvement} &\textbf{+6.43} &\textbf{+12.53} &\textbf{+12.05} &\textbf{+10.34} \\ %
      \hline
      \end{tabular}}
  \end{center}
   \caption{Comparisons for the feature-based and response-based distillation. The results are evaluated with 3D mAP on moderate  Cars, Pedestrians and Cyclists. FD and RD are short for feature-based distillation in LIGA~\cite{guo2021liga} and the response-based distillation in \ourMethod{}.}
   \label{sub_abl:cmp}
\end{table}

\section{Conclusions}
This paper presents an effective cross-modal distillation approach termed \ourMethod{} from the response levels for the stereo 3D detection task. The core designs of \ourMethod{} are the proposed X-component Guided Distillation~(XGD) for regression and the Cross-anchor Logit Distillation~(CLD) for classification. The extension ablation studies demonstrate the superiority of our proposed XGD for regression and CLD for classification.
Finally, \ourMethod{} achieves state-of-the-art performance among stereo-based detectors without introducing extra cost in the inference process compared to our stereo model on the KITTI test benchmark and the large-scale Argoverse dataset. In the future, we wish  \ourMethod{} can be applied to more 3D detectors to improve their performance.
%based on cross-modal knowledge distillation.

\section{Acknowledgement}
This work was supported by the National Science Fund of China for Distinguished Young Scholars (No. 62225603).

\bibliography{aaai23}

\end{document}